\documentclass[letterpaper]{article} 
\usepackage{aaai24}  
\usepackage{times}  
\usepackage{helvet}  
\usepackage{courier}  
\usepackage[hyphens]{url}  
\usepackage{graphicx} 
\usepackage{natbib}  
\usepackage{caption} 
\usepackage{algorithm}
\usepackage{algorithmic}
\usepackage[table,xcdraw]{xcolor}
\usepackage{graphicx}
\usepackage{amsmath}
\usepackage{amssymb}
\usepackage{booktabs}
\usepackage{bbold}
\usepackage{xcolor}
\usepackage{caption}
\usepackage{multirow}
\usepackage{soul}
\usepackage{csquotes}
\usepackage{multirow}
\usepackage{graphicx}

\usepackage{newfloat}
\usepackage{listings}
\DeclareCaptionStyle{ruled}{labelfont=normalfont,labelsep=colon,strut=off} 
\floatstyle{ruled}
\newfloat{listing}{tb}{lst}{}
\floatname{listing}{Listing}
\title{CoPL: Contextual Prompt Learning for Vision-Language Understanding}
\author{
    Koustava Goswami, Srikrishna Karanam, Prateksha Udhayanan, \\ K J Joseph and Balaji Vasan Srinivasan
}
\affiliations{
    Adobe Research, Bangalore India \\
    koustavag@adobe.com
}

\usepackage{bibentry}

\begin{document}

\maketitle

\begin{abstract}
Recent advances in multimodal learning has resulted in powerful vision-language models, whose representations are generalizable across a variety of downstream tasks. Recently, their generalization ability has been further extended by incorporating trainable prompts, borrowed from the natural language processing literature.
While such prompt learning techniques have shown impressive results, we identify that these prompts are trained based on global image features which limits itself in two aspects:
First, by using global features, these prompts could be focusing less on the discriminative foreground image, resulting in poor generalization to various out-of-distribution test cases. Second, existing work weights all prompts equally whereas intuitively, prompts should be reweighed according to the semantics of the image. 
We address these 
as part of our proposed Contextual Prompt Learning (CoPL) framework, capable of aligning the prompts to the localized features of the image. 
Our key innovations over earlier works include using local image features as part of the prompt learning process, and more crucially, learning to weight these prompts based on local features that are appropriate for the task at hand. This gives us dynamic prompts that are both aligned to local image features as well as aware of local contextual relationships. Our extensive set of experiments on a variety of standard and few-shot datasets show that our method produces substantially improved performance when compared to the current state of the art methods. We also demonstrate both few-shot and out-of-distribution performance to establish the utility of learning dynamic prompts that are aligned to local image features.
\end{abstract}

\section{Introduction}

\begin{figure*}[t]
    \centering
    \includegraphics[width= \linewidth]{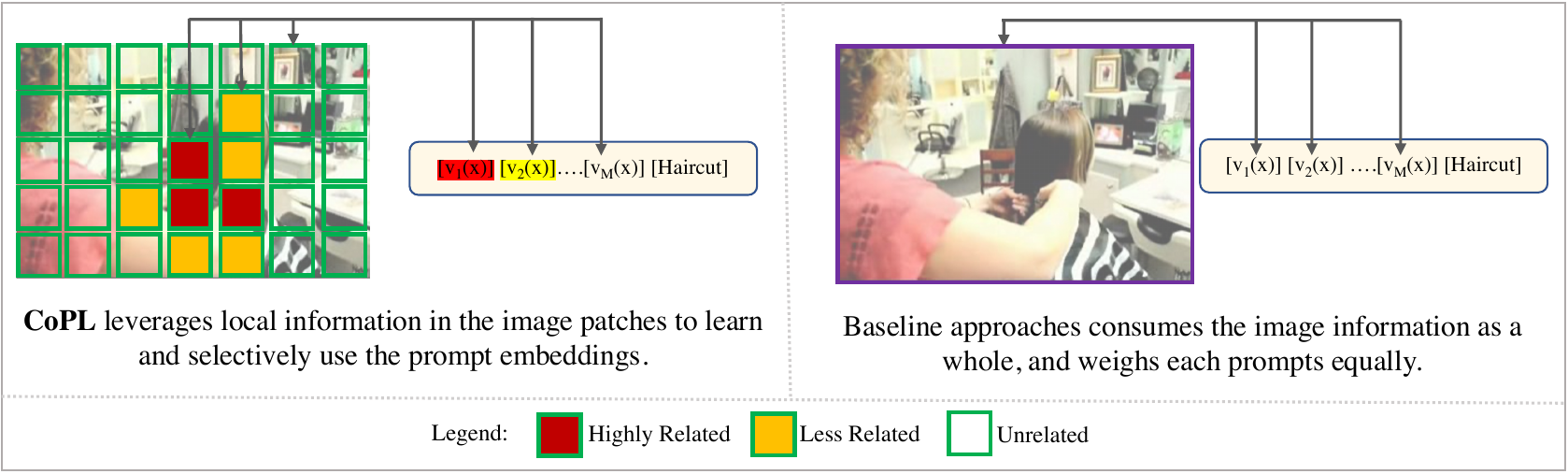}
    \caption{
    The figure summarises our methodological contribution. CoPL leverages the local information in the image patches to learn contextualized prompts (left). Further, it weights each of the prompts differently, based on its semantic affinity. The coloured patches highlights how the prompts are weighted according to their semantic alignment with the local image features. Baselines approaches (on right) uses global image features for prompt learning, and weights each of the prompts equally. 
    }
    \label{fig:CoPL_teaser_picture}
\end{figure*}


Fully supervised computer vision models for problems like classification are typically trained on datasets like ImageNet \cite{DBLP:conf/cvpr/DengDSLL009}, OpenImages \cite{DBLP:journals/corr/abs-1811-00982}, JFT300M \cite{DBLP:conf/eccv/KolesnikovBZPYG20,DBLP:conf/iccv/SunSSG17} etc., and have also proven themselves to be effective for a variety of downstream tasks via transfer learning \cite{DBLP:journals/corr/abs-2205-09904,DBLP:conf/cvpr/GuoSKGRF19,DBLP:conf/icdm/WanXLZH19}. Despite this, it is challenging to adapt these models to other domains due to various reasons including limited data and annotation overhead. Additionally, since these models are trained for specific objectives like classification, they tend to capture concepts related to categories seen during training and not to scale to unseen classes during inference. 


To enhance the adaptability of such models, 
there has been recent efforts in tuning the associated prompts (instead of the model weights). Inspired by traditional prompt engineering efforts \cite{DBLP:conf/emnlp/PetroniRRLBWM19,DBLP:conf/nips/BrownMRSKDNSSAA20,DBLP:conf/eacl/SchickS21}, there has been work in tuning discrete prompts from predefined prompt templates \cite{DBLP:conf/icml/RadfordKHRGASAM21} that help in capturing rich semantics from user intents and align them to visual contents.
However, since building a rich semantic based prompt templates require domain specific and linguistic knowledge, this apporach is not scalable. 
The CoOp \cite{DBLP:journals/ijcv/ZhouYLL22} algorithm used ideas from soft-prompting in natural language processing \cite{DBLP:conf/acl/GaoFC20,DBLP:journals/tacl/JiangXAN20,DBLP:conf/emnlp/LesterAC21,DBLP:conf/acl/LiL20,DBLP:journals/corr/abs-2103-10385,DBLP:conf/acl/LiuJFTDY022} to train dynamic learnable prompt vectors with backpropagation and preserve the semantic relationship between sentences and labels \cite{DBLP:journals/corr/abs-2103-10385}. However, the context learned with CoOp in this fashion fails to generalize to unseen classes, leading to the need for dynamically updating prompts based on the image context, an idea that was proposed in CoCoOp \cite{DBLP:conf/cvpr/ZhouYL022}. This model was trained by explicitly conditioning the prompts on image feature vectors as tokens where a separate lightweight neural network (called meta-net in their work) was used to equally weight all the prompt vectors. Later \citet{DBLP:journals/corr/abs-2303-13283} note that CoOp-based methods suffer from catastrophic knowledge forgetting, where these methods gradually miss out on essential general textual knowledge. This leads them to propose Knowledge-guided Context Optimization (KgCoOp) where they try to reduce down the distance between hand-crafted prompts and learnable prompts during training.

In our work, we identify some key issues with the aforementioned architectures. First, the features obtained from meta-net in CoCoOp \cite{DBLP:conf/cvpr/ZhouYL022} are global in nature and hence susceptible to issues like clutter and noise in many few-shot and out-of-distribution test cases (we demonstrate this empirically later on). Next, these features are directly added to all the learned prompt vectors, thus resulting in an equal weighting for each prompt. Consequently, this model is unable to learn which of the prompt vectors are more semantically relevant and contextually meaningful during inference, which makes model less generalizable  on unseen classes in a zero-shot settings. While \citet{DBLP:journals/corr/abs-2303-13283} in KgCoOp architecture prevents the catastrophic knowledge forgetting by minimizing the distance between hand-crafted prompts and learnable prompts, there is no clear evidence of what are the efficient discrete prompts. Moreover, it is hard to design generalized hand-crafted discrete prompts and sometimes tend to overfit linguistic features \cite{DBLP:journals/corr/abs-2103-10385}.


To address the aforementioned problems, we propose a new technique called \textit{Contextual Prompt Learning (CoPL)}. Our key ideas include aligning prompts to local image context, realized with local features, and determining which prompts are more sematically relevant conditioned on such local context. Our insight is that by doing so, we are able to learn a more appropriate weighting of the prompts, unlike equal weighting of prior work above, that is semantically reflective of the actual content of the image under consideration. Inspired by the concept of \textit{global attention} from the NLP literature \cite{DBLP:conf/emnlp/LuongPM15}, during training, we propose to align each local feature vector (e.g., computed from a local image patch) to a set of dynamic \textit{soft-prompts} using a learned context vector that attends to these prompt vectors. An overview of the methdology is shown in Figure \ref{fig:CoPL_teaser_picture}. 
This produces a set of attention weights for the prompt vectors that are semantically aligned to local image regions. This results in CoPL learning being more generalizable features as we demonstrate with an extensive set of zero-shot and few-shot classification evaluation settings. 

We conduct a comprehensive set of experiments on visual classification on $11$ different datasets 
and scenarios (zero-shot, one-shot, seen/unseen, and within-dataset and cross-dataset). Across all these experiments, we demonstrate substantial performance improvement when compared to all the baselines,
indicating the ability of our method to be adapted across various classification settings with little or no training and much reduced prompt engineering.



Our key contributions are: \textbf{(i)} We identify two key shortcomings with existing prompt-based image classification methods: equal weighting of the prompt vectors and no flow of contextual local information of input images to the prompt vectors while learning during back-propagation; \textbf{(ii)} We propose \textit{CoPL: Contextualized Prompt Learning}, a new method that addresses the issues above by learning prompt weights dynamically and aligning the resulting prompt vectors with local image features; \textbf{(iii)} We conduct extensive experiments with CoPL under a variety of classification scenarios and demonstrate substantial performance improvements, particularly in various unseen and few-shot data scenarios. 
Importantly, on $11$ different image recognition dataset, on average CoPL achieves state-of-the-art performance on unseen class classifications beating state-of-the-art \textbf{zero-shot large-scale model CLIP by $1.4\%$}, \textbf{conditional prompting model CoCoOp by $3.9\%$}, \textbf{knowledge pruning model KgCoOp by $2.0\%$} and \textbf{gradient learning method ProGrad by $4.9\%$} on accuracy. We evaluated CoPL on cross-dataset zero-shot image recognition tasks and on an average on $8$ datasets, CoPL outperformed CoCoOp by $2.3\%$ on accuracy. Our extensive experiments on domain generalization in Table \ref{tab:gener}, established CoPL as state-of-the-art most generalized model by outperforming MaPLe \cite{DBLP:journals/corr/abs-2210-03117} and CoPrompt \cite{DBLP:journals/corr/abs-2306-01195}. 

\section{Related Work}

\paragraph{Multimodal Models:}
Vision-language models have shown great potential in learning generic visual representations. The core idea has been to use natural language supervision for image representation learning and align them jointly in the same embedding space \cite{DBLP:conf/icml/JiaYXCPPLSLD21,DBLP:conf/icml/RadfordKHRGASAM21}.  
Early explorations in this line of research include related problem formulations -  metric learning~\cite{frome2013devise}, multilabel classification~\cite{gomez2017self, joulin2016learning}, n-gram language learning~\cite{li2017learning}, captioning~\cite{desai2021virtex}. Traditionally, hand-crafted descriptors \cite{DBLP:conf/iccv/ElhoseinySE13,DBLP:conf/nips/SocherGMN13} were the mode of capturing image representations. Later, convolutional neural net based architectures \cite{DBLP:conf/nips/FromeCSBDRM13,DBLP:conf/iccv/BaSFS15} were introduced. Recent works have focused on learning joint representations of both the modalities using deep learning architectures \cite{DBLP:journals/corr/abs-2110-11316,DBLP:conf/icml/JiaYXCPPLSLD21,DBLP:conf/iclr/LiLZCOSYY22,DBLP:conf/iccv/KamathSLSMC21}. With the introduction of transformers \cite{DBLP:conf/nips/VaswaniSPUJGKP17}, Li et al., \cite{DBLP:journals/corr/abs-1908-03557} proposed Visual-BERT, where texts and images are jointly encoded in a single transformers architecture.
One of the biggest milestone in the multimodal research is the CLIP model \cite{DBLP:conf/icml/RadfordKHRGASAM21}. It is a dual encoder based model and during training it matches pairs of images and texts. One of the main component of CLIP is the carefully designed prompts which most of the times very hard to formulate. To overcome this, Zhou et al., \cite{DBLP:journals/ijcv/ZhouYLL22} designed CoOp, which trains dynamic soft-prompts during back-propagation.

\paragraph{Prompting:} The intuition behind prompt learning is to capture user intention and instructions to perform certain downstream tasks \cite{DBLP:conf/acl/LiYL022,DBLP:conf/naacl/ParmarMPLMB22,DBLP:journals/corr/abs-2202-11345}. 
With the introduction of GPT \cite{DBLP:conf/nips/BrownMRSKDNSSAA20}, prompt engineering is shown to be performing efficiently in few-shot knowledge adaptation. 
But, building prompt templates is hard and requires immense skills. Recently, researchers have proposed \enquote{soft-prompting}. 
The main intuition is to learn dynamic continuous prompt tokens during back-propagation \cite{DBLP:conf/acl/GaoFC20,DBLP:journals/tacl/JiangXAN20,DBLP:conf/emnlp/LesterAC21,DBLP:conf/acl/LiL20,DBLP:journals/corr/abs-2103-10385,DBLP:conf/acl/LiuJFTDY022}. 
Recently, Goswami et al. \cite{DBLP:journals/corr/abs-2302-06868} 
highlights that the soft prompts can be further tuned with the semantic knowledge of language models 
without explicitly verbalizers setup. At the same \citet{DBLP:conf/pakdd/SarkarGAM23} showed tuning these soft prompts can help downstream tasks. CoOp \cite{DBLP:journals/ijcv/ZhouYLL22} is designed to train these soft prompts during training. Later  \citet{DBLP:conf/cvpr/ZhouYL022} introduced CoCoOp on top of CoOp to improve the performance by conditioning on image input. On the other hand, KgCoOp \cite{DBLP:journals/corr/abs-2303-13283} and ProGrad \cite{DBLP:journals/corr/abs-2205-14865} were proposed to improve the performance of CoOp by aligning prompts towards general knowledge. While ProGrad tries to optimize the prompts with the aligned direction,  KgCoOp minimises the distance between hand-crafted prompts and learnable prompts during training. As our methodology is directly improving the CoCoOp by infusing the local context of the image during prompt training, we take CoCoOp as our direct baseline.  Along with  CoOp and CoCoOp, KgCoOp and ProGrad methodologies are directly in line with our way of learning representations, we considered these methods as our standard baselines. Recently, MaPLe \cite{DBLP:journals/corr/abs-2210-03117} and CoPrompt \cite{DBLP:journals/corr/abs-2306-01195} introduces a new way of learning by infusing prompts to both text and image encoder. Such infusion in a multi-modal setup may affect generalization,and hence in our generalization experiment in Table \ref{tab:gener}, we compare against these approaches too. 



Here, we introduce and discuss details of our proposed method \textit{CoPL: Contextualized Prompt Learning}. Since CoPL uses the same architectural backbone as CLIP \cite{DBLP:conf/icml/RadfordKHRGASAM21}, we first begin with a brief review followed by a discussion on how our closest baseline, CoCoOp \cite{DBLP:conf/cvpr/ZhouYL022}, is trained. 
While we use the CLIP backbone for simplicity, we are in no way limited by this design choice. In fact, our method is very much applicable to be used in conjunction with a variety of other architectures, e.g., VisualBERT \cite{DBLP:journals/corr/abs-1908-03557}, MDETR \cite{DBLP:conf/iccv/KamathSLSMC21}, GLIP \cite{DBLP:conf/cvpr/LiZZYLZWYZHCG22} etc.

\subsection{Review of CLIP}

CLIP \cite{DBLP:conf/icml/RadfordKHRGASAM21} is trained using the standard contrastive learning setup where there is a text encoder and an image encoder and the overall objective function is to get their outputs as close as possible in the joint space. The text encoder is implemented with the Transformer model \cite{DBLP:conf/nips/VaswaniSPUJGKP17} which takes word sequences as input and produces both individual sequence-level as well as overall sentence-level representations. The image encoder is implemented with the ViT architecture \cite{DBLP:conf/iclr/DosovitskiyB0WZ21} which produces local (at the patch level) as well as global image features. The contrastive loss function is designed to capture the similarity between the relevant text and images, that is, the cosine similarity between the related text and image will be maximized whereas the cosine similarity with all other unrelated pairs will be minimized. To be precise, for a $K$-class classification problem, the discrete prompts are designed to have one \enquote{{class}} token. The training objective is then to fill the token with the $i^{th}$ class name where $w_i$ is the weight vector generated by the text-encoder for the same. 


\subsection{Review of CoOp/CoCoOp}
\textit{CoCoOp: Conditional Context Optimization} is built on top of a previously introduced \textit{Context Optimization (CoOp)} \cite{DBLP:journals/ijcv/ZhouYLL22} algorithm. 
It is observed that the choice of prompts plays a major role in vision-language understanding \cite{DBLP:conf/icml/RadfordKHRGASAM21}, though it is quite hard to find the best match between prompts and image descriptions. 
To overcome such a need for prompt engineering, the CoOp model \cite{DBLP:journals/ijcv/ZhouYLL22} trains a set of continuous vectors $\{\mathbf{v}_1,\mathbf{v}_2,\ldots,\mathbf{v}_M\}$ as context tokens during back-propagation. 
The CoCoOp \cite{DBLP:conf/cvpr/ZhouYL022} algorithm improves CoOp's \cite{DBLP:journals/ijcv/ZhouYLL22} performance by learning to generate prompts conditioned on each image instance, i.e., these now change for each image unlike in CoOp, where they are fixed. CoCoOp does this by training an additional two-layer neural network called meta-net that takes an image feature vector as input and produces a conditional vector that is combined with the prompt vectors to generate the final image-dependent prompts. 
This is done as follows:
\begin{align} 
     \label{meta-net} \mathbf{v}_m(\mathbf{x}) = \mathbf{v}_m + h_\theta(\mathbf{x}) 
\end{align}

where $h_\theta$ refers to the meta-net and $\mathbf{x}$ the image feature vector.

\section{Methodology}

\begin{figure}[t]
    \centering
    \includegraphics[width= \linewidth]{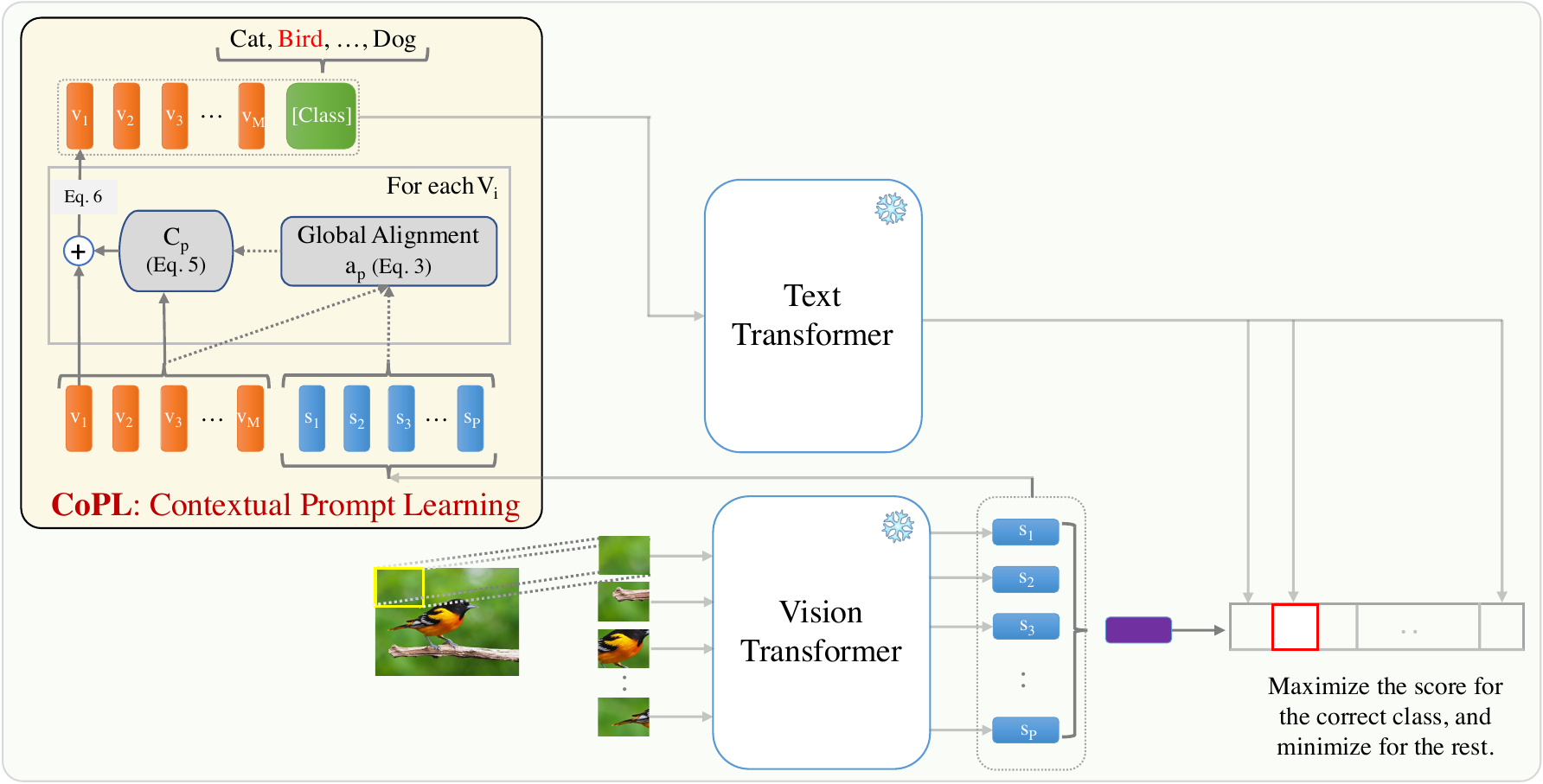}
    \caption{The overall architecture of CoPL: Contextual Prompt Learning is illustrated in this figure. It improves generalization of pretrained CLIP encoders by using contextual information from image patches, for prompt learning.}
    \label{fig:CoPL}
\end{figure}
\subsection{CoPL: Contextual Prompt Learning}
While CoCoOp generalizes better than CoOp for unseen-class classification, there is significant scope for improvement. 
First, since CoCoOp uses global feature vectors for learning the updated prompts, it focuses less on the discriminative regions in images (which tend to be more local). Next, the addition operation performed in Equation \ref{meta-net} does not capture the individual importance of each prompt token. This is particularly important since certain discriminative regions in images may weight certain prompts more when compared to others and this is not captured in the CoCoOp model. To address these issues, we propose CoPL, a simple and intuitive algorithm that operates at the local feature granularity while also aligning them with prompts to learn prompt importance weights.

Given an image $\mathbf{I}$, we first compute a set of local feature vectors. For instance, this can be the output of a vision transformer model that generates patch embeddings. Let $\mathbf{s} \in\mathbb{R}^{P\times B \times d}$ (where $P$ is number of patches from image, $B$ is training batch size, and $d$ is feature dimesionality) be this. Conditioned on these local features, we determine the semantically most meaningful prompts. To do this, we take inspiration from the attention work of Luong et al. \cite{DBLP:conf/emnlp/LuongPM15} and generate context representations that explicitly consider both the learnable prompt tokens as well as the patch representations. We first learn a lightweight neural network to generate a conditional token for the representation of each patch in $\mathbf{s}$ as:
\begin{equation}
    \mathbf{s}_p = h_\theta(\mathbf{s}_p)
\end{equation}
where $p\in \{1, 2, \ldots, P\}$. This makes the architecture parameter-efficient and easily differentiable during back-propagation. To generate the context representations that can be used to update the prompt tokens, we first learn a variable length alignment vector $\mathbf{a}_p$ ($\mathbf{a}\in\mathbb{R}^{B\times M \times d}$), one for every patch $p$ that attends to each prompt tokens $\mathbf{v}_i$ and compares them to the corresponding image patch representation $\mathbf{s}_p$ as: 

\begin{equation}
   \begin{aligned}
    \mathbf{a}_p &= \text{align} (\mathbf{s}_p,\mathbf{v}_i) \\
    &= \frac{\text{exp}(\text{score}(\mathbf{s}_p,\mathbf{v}_i))}{\sum_{i=1}^{M}(\text{exp}(\text{score}(\mathbf{s}_p,\mathbf{v}_i)))}
\end{aligned} 
\end{equation}
where $i \in {1,2,\ldots,M}$ and $M$ is the number of prompt tokens. In our setup, \texttt{score} refers to the \textit{content function} and is implemented as:

\begin{equation}
    \text{score}(\mathbf{s}_p,\mathbf{v}_i) = \text{tanh}(\mathbf{W}_a[\mathbf{s}_p;\mathbf{v}_i]) 
\end{equation}
where $W_a$ is the weight vector.

Finally, the per-patch context representation $\mathbf{c}_p$ is calculated as the weighted sum over all prompt tokens as:

\begin{equation}
    \mathbf{c}_p = \sum_{i=1}^{M}\mathbf{a}_{pi}\mathbf{v}_i
\end{equation}

The final prompt tokens are now obtained by conditioning on the context vectors above as:
\begin{equation}
    \begin{aligned}
        \mathbf{v}_m(\mathbf{x}) =  \mathbf{v}_{m}+ \sum_{i=1}^{P}\mathbf{c}_{i} 
    \end{aligned}
\end{equation}

In a nutshell, CoPL calculates the prompts for the $i$-th class as $t_i =[v_1(x),v_2(x),...,v_M(x),cl_i]$, where $cl_i$ is the embedding of the $i^{th}$ class label. The prediction probability is calculated as: 

\begin{equation}
   p(y|x) = \frac{exp(sim(x,g(t_y(x))/\gamma}{\sum_{i=1}^{k}exp(sim(x,g(t_i(x))/\gamma}
\end{equation}
where $g()$ is the feature vector produced by the text encoder and $\gamma$ is a temperature parameter. During our entire pipeline, the pre-trained CLIP model is fixed. 

\section{Experimental Evaluation} \label{sec:experiments}

\begin{table*}[]
\resizebox{\textwidth}{!}{%
\begin{tabular}{l|l|c|c|c|c|c|c|c|c|c|c|c|c}
\toprule
Methodology              & Protocols & Caltech101 & OxfordPets & StandfordCars & Flowers102 & Food101 & FGVCAircraft & SUN397 & DTD  & ImageNet & EuroSAT & UFC101 & Average \\ \midrule
\multirow{3}{*}{CLIP}    & Seen      & 96.8       & 91.1       & 62.3          & 72.0       & 90.1    & 27.1         & 69.3   & 53.2 & 72.4     & 56.4    & 70.5   & 69.3    \\  
                         & Unseen    & 94.0       & 97.2       & 74.8          & \textbf{77.8}       & 91.2    & \textbf{36.2}         & 75.3   & \textbf{59.9} & 68.l     & 64.0    & \textbf{77.5}   & 74.2    \\  
                         & HM        & 95.4       & 94.1       & 68.6          & 74.8       & 90.6    & 31.0         & 72.2   & 56.3 & 70.2     & 60.0    & 73.8   & 71.7    \\ \midrule
\multirow{3}{*}{CoOp}    & Seen      & 98.0       & 93.6       & 78.1          & \textbf{97.6}       & 88.3    & 40.4         & 80.6   & \textbf{79.4} & 76.4     & \textbf{92.1}    & \textbf{84.6}   & \textbf{82.7}    \\  
                         & Unseen    & 89.8       & 95.2       & 60.4          & 59.6       & 82.2    & 22.3         & 65.8   & 41.1 & 67.8     & 54.7    & 56.0   & 63.2    \\  
                         & HM        & 93.7       & 94.4       & 68.1          & 74.0       & 85.1    & 28.7         & 72.5   & 54.2 & 71.9     & 68.9    & 67.4   & 71.6    \\ \midrule
\multirow{3}{*}{CoCoOp}  & Seen      & 97.9       & 95.2       & 70.4          & 94.8       & 90.7    & 33.4         & 79.7   & 77.0 & 75.9     & 87.4    & 82.3   & 80.5    \\  
                         & Unseen    & 93.8       & 97.6       & 73.4          & 71.7       & 91.2    & 23.7         & 76.8   & 56.0 & 70.4     & 60.0    & 73.4   & 71.7    \\  
                         & HM        & 95.8       & 96.4       & 72.0          & 81.7       & 90.9    & 27.7         & 78.2   & \textbf{64.8} & 73.1     & 71.2    & 77.6   & 75.8    \\ \midrule
\multirow{3}{*}{KgCoOp}  & Seen      & 97.7       & 94.6       & 71.7          & 95.0       & 90.5    & 36.2         & 80.2   & 77.5 & 75.8     & 85.6    & 82.8   & 80.7    \\  
                         & Unseen    & 94.3       & 97.7       & \textbf{75.0}          & 74.7       & 91.7    & 33.5         & 76.5   & 54.9 & 69.9     & \textbf{64.3}    & 76.6   & 73.6    \\  
                         & HM        & 96.0       & 96.1       & \textbf{73.3}          & 83.6       & 91.0    & \textbf{34.8}         & 78.3   & 64.3 & 72.7     & \textbf{73.4}    & 79.6   & 77.0    \\ \midrule
\multirow{3}{*}{ProGrad} & Seen      & 98.0       & 95.0       & \textbf{77.6}          & 95.5       & 90.3    & \textbf{40.5}         & \textbf{81.2}   & 77.3 & 77.0     & 90.1    & 84.3   & 82.4    \\  
                         & Unseen    & 93.8       & 97.6       & 68.6          & 71.8       & 89.5    & 27.5         & 74.1   & 52.3 & 66.6     & 60.8    & 74.9   & 70.7    \\  
                         & HM        & 95.9       & 96.3       & 72.8          & 82.0       & 89.9    & 32.8         & 77.5   & 62.4 & 71.4     & 72.6    & 79.3   & 76.1    \\ \midrule
\multirow{3}{*}{\textbf{\textit{CoPL}}}    & Seen      & \textbf{98.1}       & \textbf{95.6}       & 70.7          & 96.1       & \textbf{90.9}    & 36.1         & 80.2   & 78.0 & \textbf{77.8}     & 89.2    & 83.1   & 81.4    \\  
                         & Unseen    & \textbf{94.9}       & \textbf{97.8}       & 74.4          & 72.1       & \textbf{91.4}    & 31.3         & \textbf{77.3}   & 50.0 & \textbf{71.3}     & 34.2    & 76.6   & \textbf{75.6}    \\  
                         & HM        & \textbf{96.5}       & \textbf{96.7}       & 72.6          & \textbf{84.1}       & \textbf{91.1}    & 33.7         & \textbf{78.7}   & 64.0 & \textbf{74.5}     & 61.7    & \textbf{79.8}   & \textbf{78.5}    \\ \bottomrule
\end{tabular}%
}
\caption{Performance of CoPL and the baselines on $11$ classification datasets; each training dataset consists $16$-shots per class. These results highlights that CoPL has better generalization than other state-of-the-art methods. The results for the base methods are borrowed from KgCoOp \cite{DBLP:journals/corr/abs-2303-13283}. HM refers to Harmonic Mean between the seen and unseen classes.}
\label{main_table}
\end{table*}


\paragraph{Datasets} We follow Zhou et al. \cite{DBLP:journals/ijcv/ZhouYLL22} to evaluate our model on $11$ image classification dataset of varying complexity. The datasets include: generic classification datasets like ImageNet \cite{deng2009imagenet} and Caltech-101 \cite{fei2004learning}; curated fine-grained datasets like OxfordPets \cite{DBLP:conf/cvpr/ParkhiVZJ12}, StanfordCars \cite{DBLP:conf/iccvw/Krause0DF13}, Flowers102 \cite{DBLP:conf/icvgip/NilsbackZ08}, Food101 \cite{DBLP:conf/eccv/BossardGG14} and FGVCAircraft \cite{DBLP:journals/corr/MajiRKBV13}; scene, action, texture and satellite image recognition datasets from SUN397 \cite{DBLP:conf/cvpr/XiaoHEOT10},  UCF101 \cite{DBLP:journals/corr/abs-1212-0402},  DTD \cite{DBLP:conf/cvpr/CimpoiMKMV14} and EuroSat \cite{DBLP:conf/igarss/HelberBDB18} respectively. For the few-shot experiments, we follow Zhou et al. \cite{DBLP:conf/cvpr/ZhouYL022} to randomly sample datapoints for training. The models are evaluated on the entire test set to report the accuracy. 

\paragraph{Training Details} 

In CLIP architecture, we use  ViT-B/$16$ as image encoder. Our prompt token length is $4$. All our models are trained with a batch size $1$ for $10$ epochs on a single $16$ GB Tesla T$4$ GPU system.
Our starting learning rate is $0.002$ and used cosine learning rate scheduler. Our warm-up with a  constant learning rate is $0.00001$.

\paragraph{Baseline Models} We  compare our approach with CoCoOp \cite{DBLP:conf/cvpr/ZhouYL022}, CoOp \cite{DBLP:journals/ijcv/ZhouYLL22}, KgCoOp \cite{DBLP:journals/corr/abs-2303-13283}, ProGrad \cite{DBLP:journals/corr/abs-2205-14865} and also large-scale zero-shot methodology CLIP \cite{DBLP:conf/icml/RadfordKHRGASAM21}. 
While comparing with CLIP, we indeed compare our learned prompt embeddings with manually designed prompts. We closely follow most of our experimental setting with that of Zhou et al., \cite{DBLP:conf/cvpr/ZhouYL022}.


\begin{table*}[t] \centering
\small
\begin{tabular}{lcccccccccc}
\toprule
\multicolumn{1}{l}{Methodology} & \multicolumn{1}{l}{\begin{tabular}[c]{@{}l@{}}Oxford\\ Pets\end{tabular}} & \multicolumn{1}{l}{\begin{tabular}[c]{@{}l@{}}Standford\\ Cars\end{tabular}} & \multicolumn{1}{l}{Food101} & \multicolumn{1}{l}{DTD} & \multicolumn{1}{l}{EuroSat} & \multicolumn{1}{l}{Flowers102} & \multicolumn{1}{l}{\begin{tabular}[c]{@{}l@{}}FGVC\\ Aircraft\end{tabular}} & \multicolumn{1}{l}{UCF101} & \multicolumn{1}{l}{SUN397} &  \multicolumn{1}{l}{Avg} \\ \midrule
CoCoOp                            & 89.6                                                                       & 61.4                                                                          & 88.2                         & 47.0                     & \textbf{63.5}                & 66.6                            & 20.9                                                                         & 65.5                        &       70.3      &    63.7            \\
CoPL                              & \textbf{89.8}                                                              & \textbf{61.9}                                                                 & \textbf{89.5}                & \textbf{49.5}            & 58.1                         & \textbf{71.3}                   & \textbf{23.8}                                                                & \textbf{70.3}               &  \textbf{71.9} & \textbf{65.1} \\ \bottomrule                         
\end{tabular}
\caption{Table shows the zero-shot image recognition results of CoPL across multiple dataset. We trained models on Caltech-101 dataset and tested on the other datasets. These results suggests very strong generalization ability of CoPL.}
\label{zero-shot}
\end{table*}

\begin{table}[]
\resizebox{0.48\textwidth}{!}{%
\begin{tabular}{@{}l|c|ccccc@{}}
\toprule
\multirow{2}{*}{Models} & Source        & \multicolumn{5}{c}{Target}                                                      \\ \cmidrule(l){2-7} 
                        & ImageNet      & ImageNetV2    & ImageNet-Sk & ImageNet-A    & ImageNet-R    & Average       \\ \cmidrule(r){1-7}
CLIP                    & 66.7          & 60.8          & 46.1            & 47.7          & 73.9          & 57.1          \\
CoCoOp                  & 71.0          & 64.0          & 48.7            & 50.6          & 76.1          & 59.9          \\
CoOp                    & 71.5          & 64.2          & 47.9            & 49.7          & 75.2          & 59.2          \\
ProGrad                 & 72.2          & 64.7          & 47.6            & 49.3          & 74.5          & 59.0          \\
KgCoOp                  & 71.2          & 64.1          & 48.9            & 50.6          & 76.7          & 60.1          \\
MaPLe                   & 70.7          & 64.0          & 49.1            & \textbf{50.9} & 76.9          & 60.2          \\
CoPrompt                   & 70.8          & 64.2          & 49.4            & 50.5 & 77.5          & 60.4          \\ \midrule
\textbf{\textit{CoPL}}                    & \textbf{71.7} & \textbf{65.2} & \textbf{49.4}   & 50.8          & \textbf{78.6} & \textbf{61.0} \\ \bottomrule
\end{tabular}%
}
\caption{Comparison of CoPL with existing approaches in domain generalization setting. On an average, CoPL outperforms other baseline results.}
\label{tab:gener}
\end{table}

\subsection{Knowledge Transfer to Unseen Classes}
One of the main downsides of CoOp is that it is unable to generalize to the unseen classes while being trained on the base classes. 
The other methods improve over CoOp, but they miss out on capturing the local features of the image and do not give weightage to the semantically aligned learnable prompts, which is the key focus area of our work. Following the implementation of Zhou et al., \cite{DBLP:conf/cvpr/ZhouYL022}, we conducted our experiments on the above mentioned $11$ datasets both for seen and unseen classes. While training is only conducted on the base classes, during testing we transfer the learnt knowledge to classify unseen classes as well as seen classes. 

From Table \ref{main_table}, we observe that there is a decrease in performance for CoOp methodology over the zero-shot large-scale method CLIP for unseen classes. Though introduction of conditional prompting improve the performance of CoCoOp, in most of the cases it fails to generalize to unseen classes with the dynamically learnt prompts (in $7$ out of $11$ datasets CLIP outperform CoCoOp methodology). In fact KgCoOp, though it is not conditioned on the image features during prompt learning, outperforms CoCoOp in many cases. 

Interestingly, our proposed approach CoPL, improves the accuracy of unseen classes over CoCoOp and CoOp for $10$ different tasks. It also outperforms KgCoOp and ProGrad during unseen class classification in $6$ tasks.  Moreover, the harmonic mean for all $7$ tasks are greater than the current state-of-the-art model, which specifies that the learned prompts are more generalizable across domains and tasks. It is important to note that CoPL comfortably outperforms manual prompt based methodology CLIP on $5$ different tasks. With the relatively complex dataset like EuroSAT, where localization is hard to be aligned due to the nature of satellite imagery, for the seen classes CoPL outperformed CoCoOp by $1.8\%$, KgCoOp by $3.6\%$ and CLIP by $32.8\%$ on accuracy.

\begin{figure}[ht!]
    \centering
    \includegraphics[width= \linewidth] {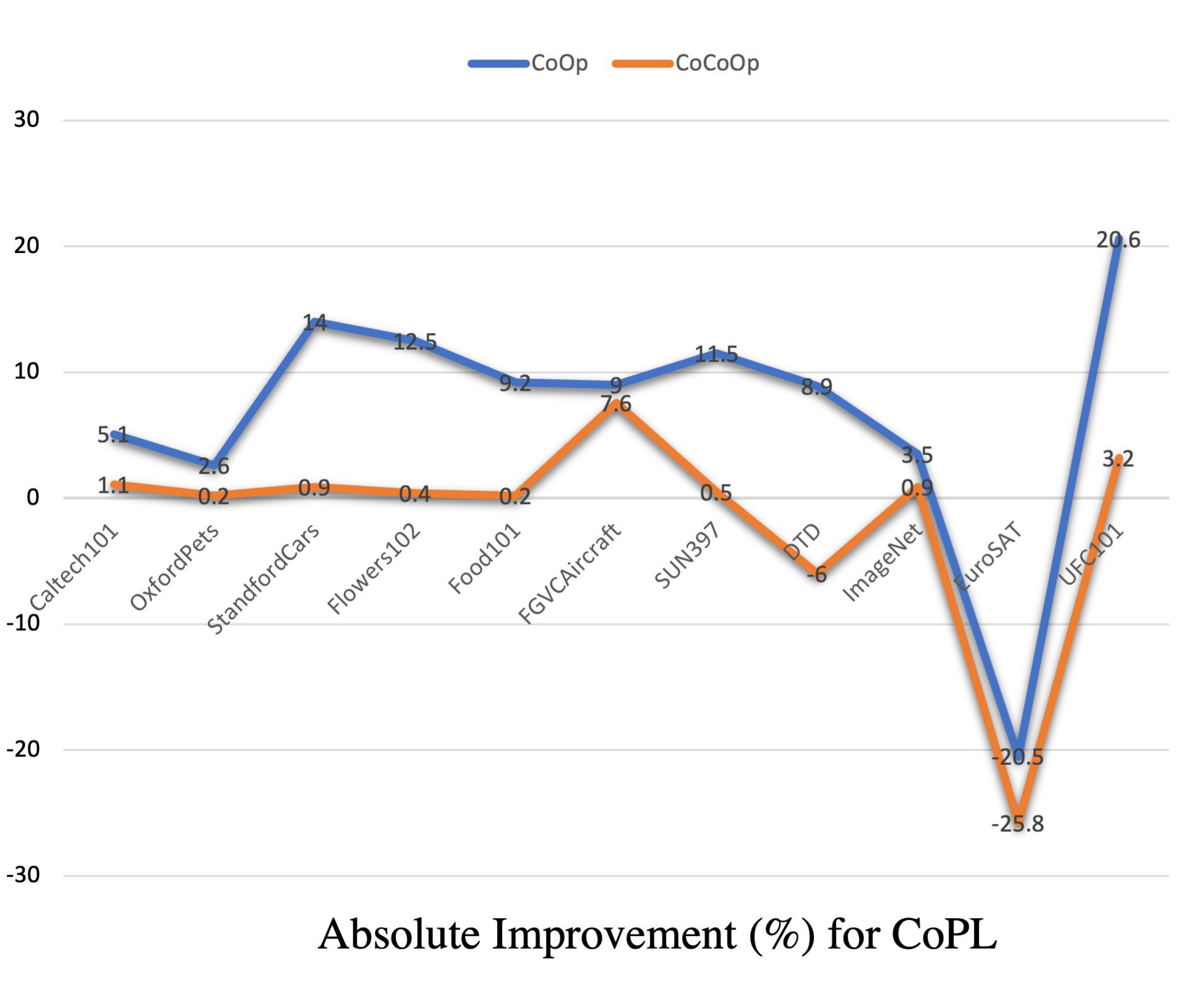}
    \caption{Performance comparison of CoPL with CoOp and CoCoOp for the unseen class detection while trained on base classes within the dataset.}
    \label{fig:CopLvsCoOp}
\end{figure}
In Figure \ref{fig:CopLvsCoOp}, we further analyse the absolute performance gain on the unseen classes over CoOp and CoCoOp methodologies. When compared with CoOp on UFC101 dataset, CoPL improve the accuracy by $20.6\%$, showing the capability of the dynamic prompt vectors to learn semantic relatedness conditioned on the local image features. On FGVC-Aircraft dataset, the absolute performance gain of $9.0\%$ and $7.6\%$ for CoOp and CoCoOp respectively. Performance boost of $8.9\%$ on the DTD dataset, confirms the capability of COPL in identifying different textures, opening up the possibility to deploy the model for design understanding. CoPL outperforms CLIP by $2.7\%$ on the FGVC-Aircraft dataset, bringing out its ability to transfer the learnt knowledge to the relatively complex unseen classes.

In general, for both seen and unseen dataset, CoPL outperforms \textbf{CoCoOp by $2.7\%$}, \textbf{KgCoOp by $1.5\%$} and \textbf{ProGrad by $2.4\%$} in accuracy. Also, it achieves a gain of $6.8\%$ in accuracy over manual prompt base methodology CLIP, suggesting that conditional alignment of prompts to local features of the image can generalize the model better to diverse recognition tasks. Ideally, the manually designed prompts should have been well aligned with the images, thus CLIP can be considered as the best performing model. It is interesting to note that CoPL, on an average across all $11$ datasets, achieves state-of-the-art performance by outperforming CLIP by $1.4\%$ in accuracy. This justifies our assertion of making dynamic prompt vectors more contextualized like human annotated prompts by observing the different local features of the image.

\subsection{Inter Dataset Zero-shot Performance}

Here, we test the mettle of our approach on an even more challenging setting. We learn the model on a dataset, and evaluate the model on a different dataset to see how transferable the learned representations are. 
Concretely, we train the model on 
Caltech101 dataset and evaluate its performance on the rest of the datasets. 
Observe, in Table \ref{zero-shot}, CoPL outperforms CoCoOp across most of the datasets. For complex dataset like FGVCAircraft, CoPL  gains $2.9\%$ 
Caltech101 being a general purpose object classification dataset, we observe impressive performance on OxfordPets and Food101 dataset. Interestingly, though trained on object classification dataset, CoPL is able to transfer the knowledge to do texture recognition task on DTD dataset, outperforming CoCoOp by $2.5\%$ on accuracy. In the most challenging EuroSAT dataset, consists of images taken from satellite, CoPL achieves $58.1\%$ accuracy, comparable to the performance of the CoCoOp. These highlights that CoPL able to learn the semantics of the images by identifying the localized features and align it with the learnable textual prompts. 

\subsection{Domain Generalization}

Domain generalization matrix establishes the claim of a model being more adaptable to target dataset from same class but having different data distribution compared to sourced domain. Here with the existing baseline we have also compared our method with MaPLe \cite{DBLP:journals/corr/abs-2210-03117} and CoPrompt \cite{DBLP:journals/corr/abs-2306-01195}. We trained the model on Imagenet and tested the model on  ImageNetV2, ImageNet-Sketch, ImageNet-A, and ImageNet-R. Observe, in Table \ref{tab:gener}, CoPL on an average outperforms all the baseline methodologies. This indicates that, by learning to focus on local image features, CoPL able to figure out the subjective image-prompt mapping, which makes CoPL more robust while working on out-of-distribution dataset.

\subsection{One-shot Training}

We evaluate our approach in an extremely low data regime. 
Here, we train CoPL and CoCoOp with $1$ training instance per class for each of the image recognition task and test the accuracy on both seen and unseen classes within the dataset. In our experiments, we valuate across $7$ datasets, and discuss their performance on the seen and unseen classes next.

\begin{figure}[ht!]
    \centering
    \includegraphics[width= \linewidth] {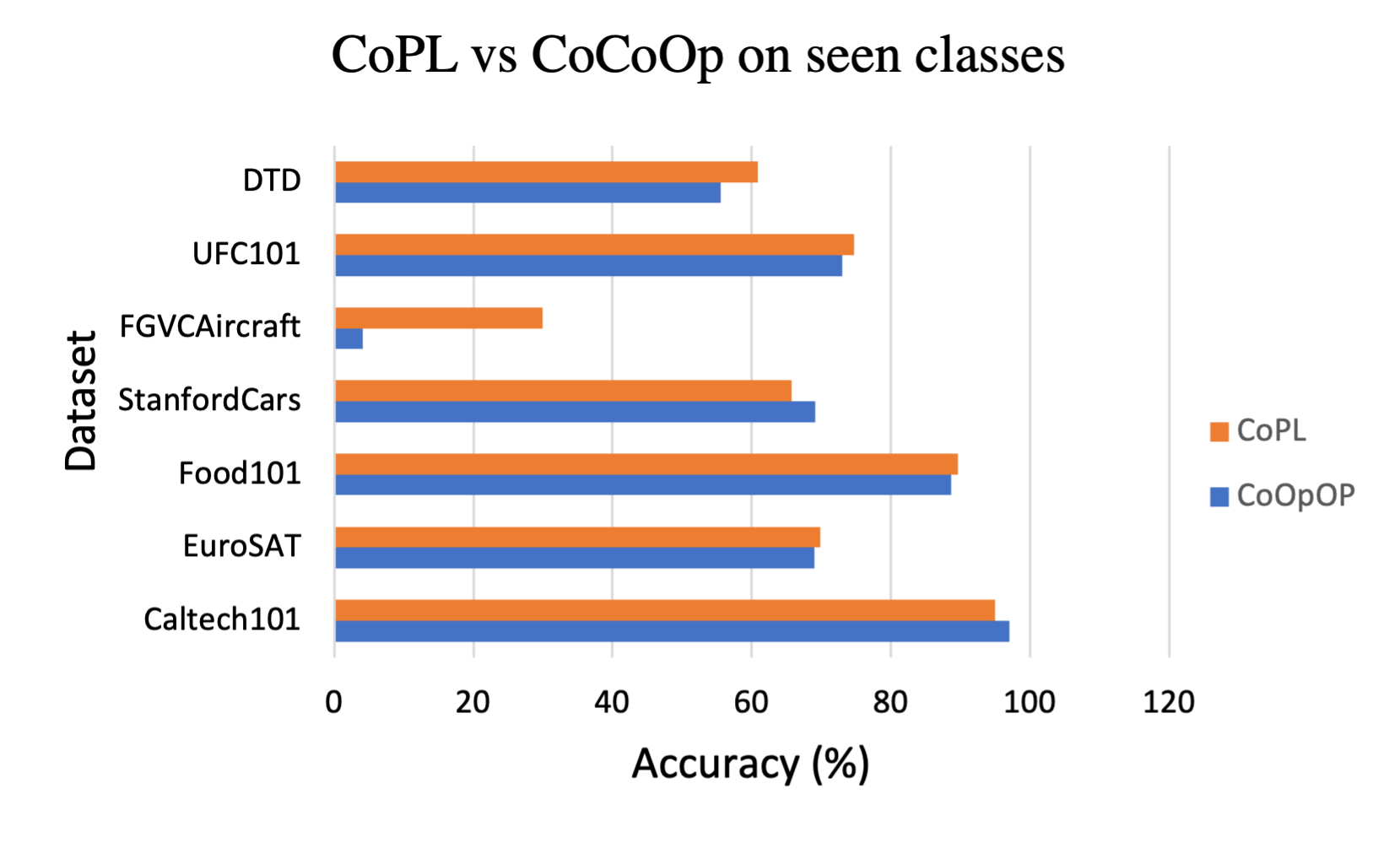}
    \caption{Comparison of CoPL with CoCoOp for the seen class classification while trained on $1$-shot setting.}
    \label{fig:seenn_class_graph}
\end{figure}

\paragraph{Seen Class} In Figure \ref{fig:seenn_class_graph}, we observe that for most of the datasets, CoPL outperforms CoCoOp indicating the capacity of adapting to the in-training datasets. It is interesting to see that, for FGVCAircraft dataset, CoPL outperforms CoCoOp by large margin ($25.9\%$ on accuracy). 

\begin{figure}[ht!]
    \centering
    \includegraphics[width= \linewidth] {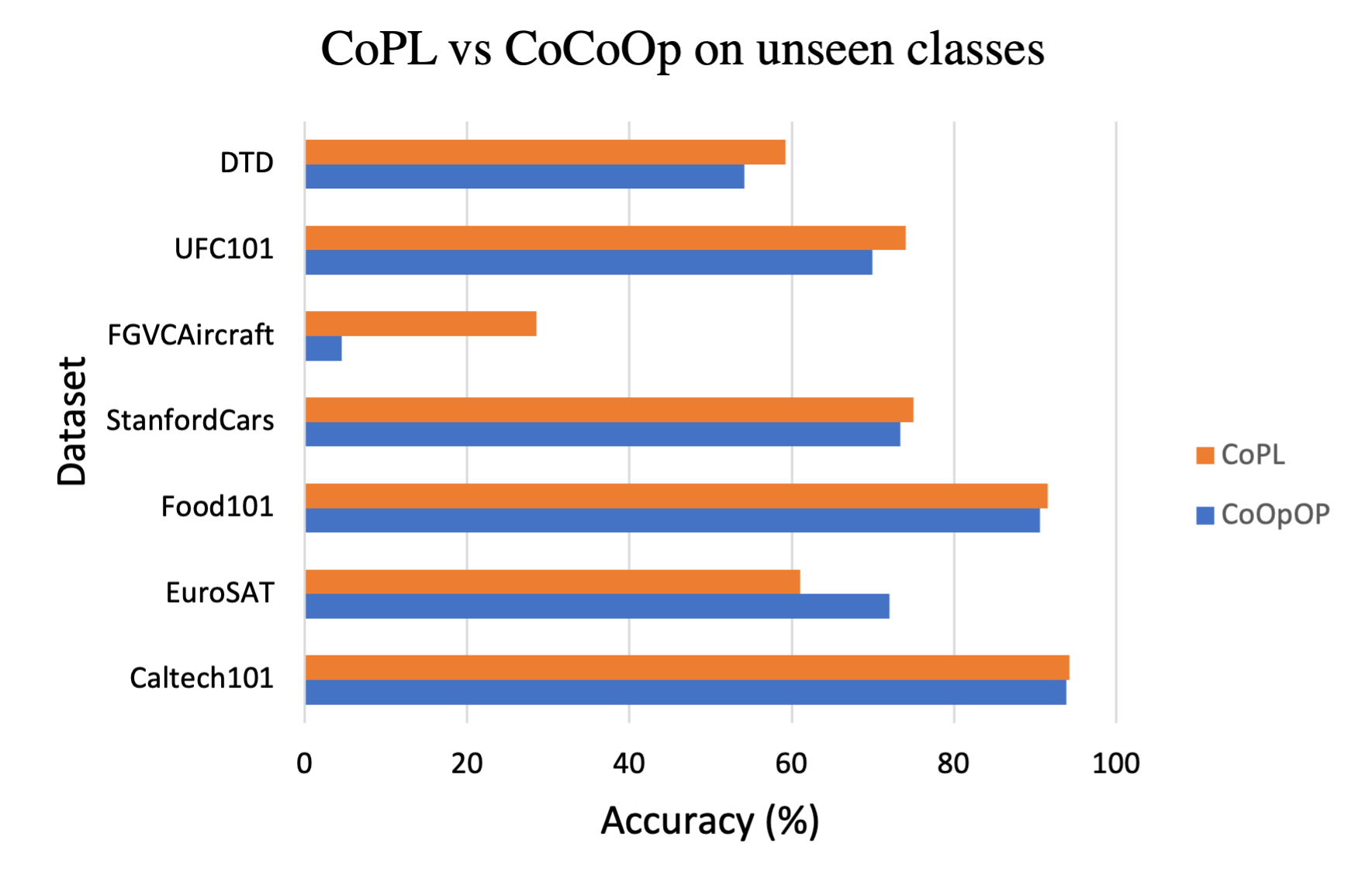}
    \caption{Comparison of CoPL with CoCoOp for the unseen classes classification while trained on $1$-shot setting.}
    \label{fig:unseen_class_graph}
\end{figure}

\paragraph{Unseen Class} The performance of the CoPL on unseen classes while only trained on $1$-shot seen classes is showcased in Figure \ref{fig:unseen_class_graph}. CoPL outperforms CoCoOp on $6$ datasets. It is interesting to observe that, on FGVCAircraft dataset CoPL improves the accuracy from $4.9\%$ to $28.5\%$. 

These results indicates that, the alignment of local image features to prompts helps to capture the semantic meaning of the images while only trained with one training instance per class. This makes the model easily adaptable to diverse set of image recognition task, even for low-resource scenarios.

\section{Discussions and Analysis}

\paragraph{Local vs Global Image Features} A key contribution of CoPL is to utilize the localized image features for prompt learning. 
Further, we give weightage to the prompts based on the semantic relatedness. 
In this section, we critically analyse the contribution of using local features an opposed to global features. 
To evaluate this, we have conduct  experiments on Caltech101 and DTD datasets. 
We selected these two dataset as they are primarily targeted for two different 
recognition tasks: object recognition and texture recognition. 
Similar to our earlier evaluation protocol, we evaluate the performance on the seen classes and unseen classes separately. From Table \ref{localvsglobal}, we can can clearly understand that for both seen and unseen classes, aligning local contextual features with prompts, helps CoPL to generalize better.

\begin{table}[]
\resizebox{0.48\textwidth}{!}{%
\begin{tabular}{@{}l|l|cc@{}}
\toprule
Methodology                                         & Class Type & Caltech101    & DTD           \\ \midrule
\multirow{2}{*}{Global Attention + Global Features} & Seen       & 97.2          & 77.4          \\
                                                    & Unseen     & 94.1          & 39.9          \\ \midrule
\multirow{2}{*}{Global Attention + Local Features}  & Seen       & \textbf{98.1} & \textbf{78.0} \\
                                                    & Unseen     & \textbf{94.9} & \textbf{50.0} \\ \bottomrule
\end{tabular}%
}
\caption{Ablation Study: importance of aligning the local contextual image features with the prompt tokens.}
\label{localvsglobal}
\end{table}

\paragraph{Incremental Test} Following the experimental analysis presented in CoCoOp \cite{DBLP:conf/cvpr/ZhouYL022}, we evaluate the model efficacy where test dataset consists of both seen and unseen classes. In this case, during training the model weights are updated based on the seen classes but has not considered the unseen classes. Thus, during testing model will be performing zero-shot classification on unseen classes. 
\begin{table}[ht!]
\centering
\begin{tabular}{l|cccc}
\toprule
Methodology & CLIP & CoOp & CoCoOp & CoPL \\ \midrule
Accuracy & 65.2 & 65.6 & 69.1 & \textbf{74.7} \\ \bottomrule
\end{tabular}
\caption{Average accuracy on $11$ datasets, when the test set contains both the seen and unseen classes. The models are trained on training instances for the seen classes. CoPL optimally balances stability and plasticity.
}
\label{test}
\end{table}
We observe in Table \ref{test}, that CoPL comfortably outperforms all the baseline models, suggesting that local image feature alignment with prompts help to generalize better. 

\paragraph{Run-time Analysis}

We experimentally analyse and quantify the extra wall-clock time that is required for CoPL when compared to the closest baseline CoCoOp. We use Caltech101 for this experiment. 
For training $800$ data-points (corresponding to $50$ classes) , CoPL takes $28$ minutes whereas evaluation on $1549$  data-points took $2$ minutes $05$ seconds. On the other hand CoCoOp takes $21$ minutes $35$ seconds for training and $1$ minute $34$ second for inference. 
From this, we understand that CoPL improves generalization without sacrificing too much on the computational overhead.  

\begin{figure}[h]
    \centering
    \includegraphics[width= \linewidth]{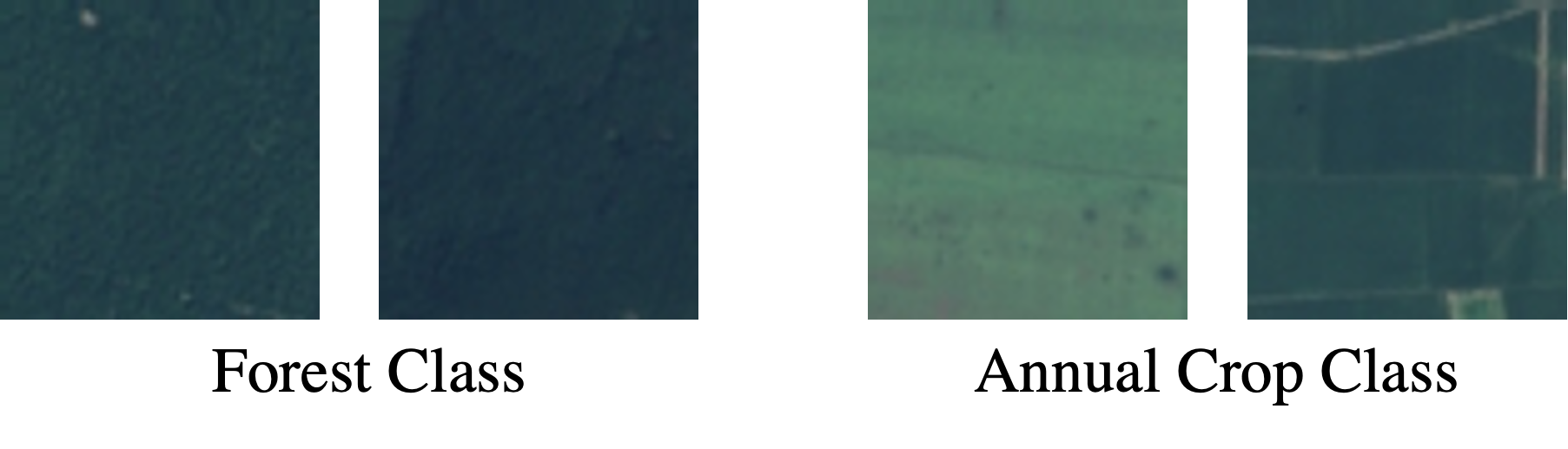}
    \caption{Visualization of samples from EuroSat dataset. We note that majority of the sampled does not contain any salient objects, which makes local feature less effective.}
    \label{fig:EuroSat}
\end{figure}

\paragraph{Limitation}
Our exhaustive experimental analysis across $11$ datasets greatly helped us to test the mettle of our method. Our performance on the EuroSat dataset in the unseen class category, is lower than CoCoOp. While analysing the reasons for the drop in performance, we could uncover that images from EuroSat does not contain images with salient objects. 
We visualise some such examples in Fig.~\ref{fig:EuroSat}; it can be seen that there are no ``local'' regions in these images, which can contribute towards modelling the prompt better.
This results in lower performance of CoPL on such datasets.

\section{Conclusion}
In this paper, we present \textit{CoPL: Contextual Prompt Learning}, which can align prompts to the corresponding contextual local image features. During alignment, we also produces a set of attention weights for the prompt vectors, that are semantically related to local image regions. Extensive experimental evaluation on $11$ image recognition datasets showcases the efficacy of CoPL in understanding the semantic relationship between the images and the prompts. Moreover, the state-of-the-art zero-shot and few-shot results justify our claim of making CoPL better in generalization by aligning the local features of the images to prompts. 
In future, we plan to make CoPL capable of understanding user intents to make local edits on images. 

\bibliography{aaai24}

\end{document}